\documentclass[runningheads]{config/llncs}

\usepackage{config/eccv}

\usepackage{config/eccvabbrv}

\usepackage{tikz}
\usepackage{booktabs}       % professional-quality tables
\usepackage{enumitem}
\usepackage{multirow}
\usepackage{adjustbox}
\usepackage{fontawesome}
\usepackage{makecell}
\usepackage{pifont}% http://ctan.org/pkg/pifont % ding % xmark
\usepackage[pagebackref=true,breaklinks=true,colorlinks,bookmarks=false]{hyperref}

\usepackage[accsupp]{axessibility}  % Improves PDF readability for those with disabilities.

\usepackage{lipsum}

\usepackage[normalem]{ulem}

\usepackage{xspace}
\usepackage{currfile}
\xspaceaddexceptions{]\}}

\newcommand{\camready}[1]{\textcolor{black}{{#1}}\xspace}

\newcommand{\nameMethod}{\mbox{HUMOS}\xspace}

\newcommand{\projectURL}{\url{https://github.com/CarstenEpic/humos}}
\definecolor{myBlue}{rgb}{0.18, 0.38, 0.82}  % \definecolor{myBlue}{rgb}{47, 97, 207}\//

\newcommand{\outTitleFULL}{HUMOS: Human Motion Model Conditioned on Body Shape}

\newcommand{\supmat}{Sup.~Mat.\xspace}

\usepackage{lipsum}

\newcommand{\smpl}{\mbox{SMPL}\xspace}

\newcommand{\colorRef}[1]{#1} % <-----
\usepackage[capitalize]{cleveref}
\crefname{figure}{\colorRef{Fig.}}{\colorRef{Figs.}}
\Crefname{figure}{\colorRef{Figure}}{\colorRef{Figures}}
\crefname{section}{\colorRef{Sec.}}{\colorRef{Secs.}}
\Crefname{section}{\colorRef{Section}}{\colorRef{Sections}}
\Crefname{table}{\colorRef{Table}}{\colorRef{Tables}}
\crefname{table}{\colorRef{Tab.}}{\colorRef{Tabs.}}
\Crefname{equation}{\colorRef{Equation}}{\colorRef{Equations}}
\crefname{equation}{\colorRef{Eq.}}{\colorRef{Eqs.}}

\newcommand{\idA}{\mathcal{A}}
\newcommand{\idB}{\mathcal{B}}
\newcommand{\gender}{\mathcal{G}}
\newcommand{\trans}{\mathbf{x}}
\newcommand{\rot}{\mathbf{r}}
\newcommand{\identity}{\mathcal{I}}

\newcommand{\volume}{\mathcal{V}}
\newcommand{\meshSurfPointi}{v_i}

\newcommand{\meshPart}{P}
\newcommand{\meshParti}{\meshPart_i}
\newcommand{\meshPartvi}{\meshPart_{\meshSurfPointi}}

\newcommand{\volumeParti}{\volume^{\meshParti}}
\newcommand{\volumePartvi}{\volume^{\meshPartvi}}

\definecolor{lightgray}{gray}{0.97}
\definecolor{lightblue}{rgb}{0.93,0.95,1.0}

\definecolor{GreenColor}{rgb}{0.137,0.573,0.565}
\definecolor{OrangeColor}{rgb}{0.914,0.541,0.0.141}
\definecolor{PurpleColor}{rgb}{0.5,0,0.7}
\definecolor{BlueColor}{rgb}{0.255,0.412,0.882}

\newlist{todolist}{itemize}{2}
\setlist[todolist]{label=$\square$}

\newcommand{\Tr}{\mathrm{Tr}}

\begin{document}

\title{\outTitleFULL}
\author{
Shashank Tripathi$^{12}$\thanks{work done during an internship at Epic Games} \quad 
Omid Taheri$^{1}$ \quad 
Christoph Lassner$^2$ \\
Michael J. Black$^1$ \quad 
Daniel Holden$^2$ \quad 
Carsten Stoll$^2$}
\institute{$^1$Max Planck Institute for Intelligent Systems, T{\"u}bingen, Germany \quad
$^2$Epic Games
% {\small
% $^1$Max Planck Institute for Intelligent Systems, T{\"u}bingen, Germany \quad
% $^2$Epic Games
% }\\
{\tt\small \{stripathi, otaheri, black\}@tue.mpg.de \\ \{christoph.lassner, daniel.holden, carsten.stoll\}@epicgames.com}
}

% DO NOT ENTER AUTHOR INFORMATION FOR ANONYMOUS TECHNICAL PAPER SUBMISSIONS TO SIGGRAPH 2019!
%\author{Gang Zhou}
%\orcid{1234-5678-9012-3456}
%\affiliation{%
%  \institution{College of William and Mary}
%  \streetaddress{104 Jamestown Rd}
%  \city{Williamsburg}
%  \state{VA}
%  \postcode{23185}
%  \country{USA}}
%\email{gang_zhou@wm.edu}
%\author{Valerie B\'eranger}
%\affiliation{%
%  \institution{Inria Paris-Rocquencourt}
%  \city{Rocquencourt}
%  \country{France}
%}
%\email{beranger@inria.fr}
%\author{Aparna Patel}
%\affiliation{%
% \institution{Rajiv Gandhi University}
% \streetaddress{Rono-Hills}
% \city{Doimukh}
% \state{Arunachal Pradesh}
% \country{India}}
%\email{aprna_patel@rguhs.ac.in}
%\author{Huifen Chan}
%\affiliation{%
%  \institution{Tsinghua University}
%  \streetaddress{30 Shuangqing Rd}
%  \city{Haidian Qu}
%  \state{Beijing Shi}
%  \country{China}
%}
%\email{chan0345@tsinghua.edu.cn}
%\author{Ting Yan}
%\affiliation{%
%  \institution{Eaton Innovation Center}
%  \city{Prague}
%  \country{Czech Republic}}
%\email{yanting02@gmail.com}
%\author{Tian He}
%\affiliation{%
%  \institution{University of Virginia}
%  \department{School of Engineering}
%  \city{Charlottesville}
%  \state{VA}
%  \postcode{22903}
%  \country{USA}
%}
%\affiliation{%
%  \institution{University of Minnesota}
%  \country{USA}}
%\email{tinghe@uva.edu}
%\author{Chengdu Huang}
%\author{John A. Stankovic}
%\author{Tarek F. Abdelzaher}
%\affiliation{%
%  \institution{University of Virginia}
%  \department{School of Engineering}
%  \city{Charlottesville}
%  \state{VA}
%  \postcode{22903}
%  \country{USA}
%}

%\renewcommand\shortauthors{Zhou, G. et al}
\authorrunning{S. Tripathi et al.}

\maketitle

\begin{center}
    \centering
    \includegraphics[width=\linewidth]{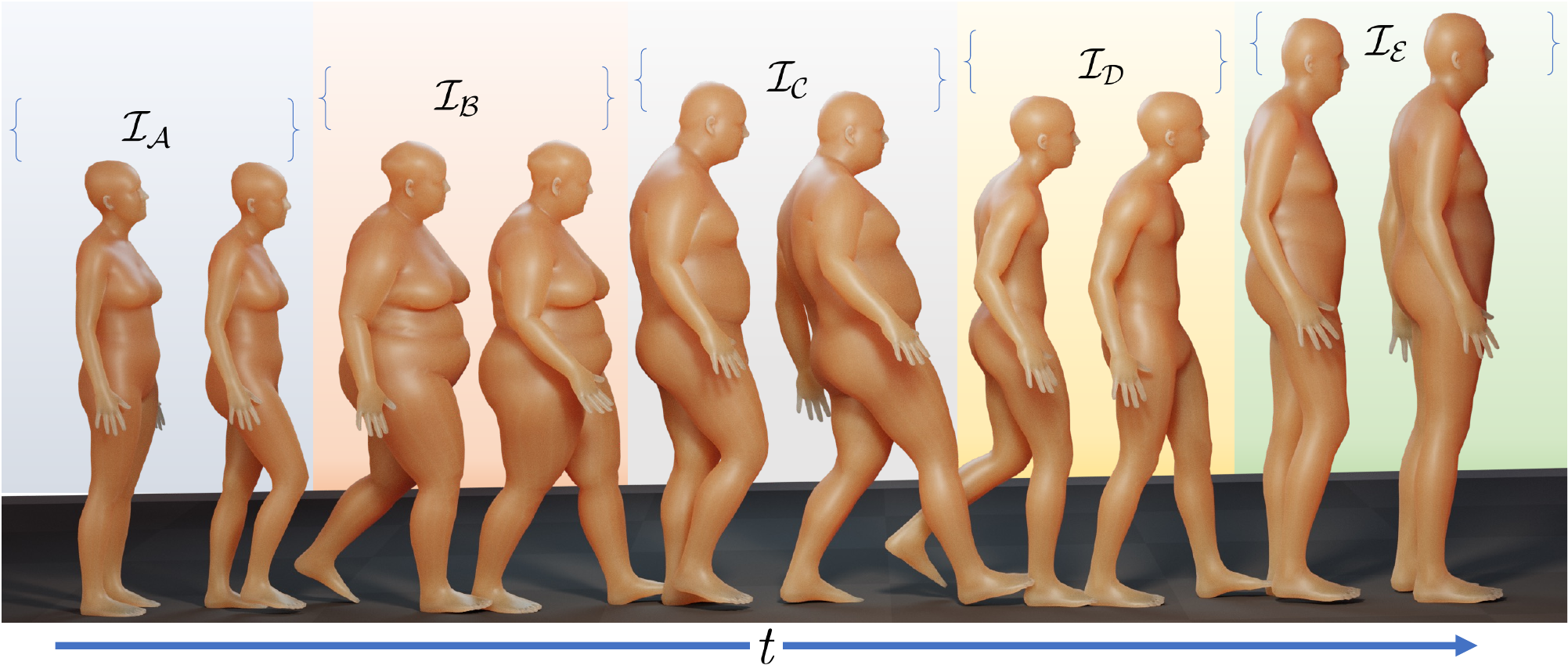}
    \captionsetup{type=figure}
    \vspace{-1em}
    \caption{People with different body shapes perform the same motion differently. Our method, \nameMethod, generates \emph{natural}, \emph{physically plausible} and \emph{dynamically stable} human motions conditioned on body shape. \nameMethod uses a novel identity-preserving cycle consistency loss and differentiable dynamic stability and physics terms to learn an identity-conditioned manifold of human motions. \camready{Shown here is the same walk motion with a skip-step in the middle, generated by \nameMethod for five different identities $\mathcal{I}_{\idA:\mathcal{E}}$. To demonstrate shape-conditioning, we visualize the same motion but successively change the identity after every 30 frames.}}
    \label{figure:tipman:teaser}
\end{center}%

\begin{abstract}
\camready{Generating realistic human motion is crucial for many computer vision and graphics applications. The rich diversity of human body shapes and sizes significantly influences how people move. However, existing motion models typically overlook these differences, using a normalized, average body instead. This results in a homogenization of motion across human bodies, with motions not aligning with their physical attributes, thus limiting diversity. To address this, we propose a novel approach to learn a generative motion model conditioned on body shape. We demonstrate that it is possible to learn such a model from unpaired training data using cycle consistency, intuitive physics, and stability constraints that model the correlation between identity and movement. The resulting model produces diverse, physically plausible, and dynamically stable human motions that are quantitatively and qualitatively more realistic than existing state of the art.}
More details are available on our project page \projectURL.

\end{abstract}

\section{Introduction}
\label{sec:intro}
Modeling virtual humans that move and interact realistically with 3D environments is extremely important for interactive entertainment, AR/VR and simulation technology, with numerous applications in crowd simulation, gaming and robotics. 
There has been rapid progress in training models that generate human motion either unconditionally or conditioned on text or previous motions.
Existing state-of-the-art human motion models~\cite{petrovich2022temos, petrovich2021actor, tevet2023mdm, chen2023latent} are trained on datasets like AMASS \cite{AMASS_2019}, but they typically ignore body shape and proportions.
However, variations in muscle mass distribution and body proportions contribute to a person's distinct movement patterns. People with different body types will generally move differently when prompted to perform the same motion. 
We argue that to achieve physical realism and motion diversity, it is critical to condition generated human motions on body shape.

To address this problem we adopt a novel approach called \nameMethod, that enhances traditional data-driven motion generation methods and uses a transformer-based conditional Variational Auto-Encoder (c-VAE) trained to generate human motion conditioned on \emph{identity} features such as a subject's body shape and sex. We take inspiration from a recent 3D human pose and shape estimation method, IPMAN~\cite{tripathi2023ipman}, to propose new dynamic intuitive physics (IP) terms that are simple, fully differentiable, and compatible with parametric body models like SMPL~\cite{SMPL:2015}.
Since IPMAN's IP terms only apply to static 3D poses, they are not suited for dynamic human motion modeling. We go beyond this by proposing general IP terms that are effective for dynamic human motions involving a sequence of poses. 
We show that these dynamic IP terms are critical to effectively train our model without paired data of differently-shaped people performing the same action.

Specifically, we propose differentiable physics terms that improve the realism of generated motions by addressing common issues like foot sliding, ground penetration, and unrealistic floating effects. Our key contribution here is a dynamic stability term, that models %
the interaction between a body's Center of Mass (CoM), Center of Pressure (CoP), and the Zero Moment Point (ZMP). Dynamic stability is a biomechanical concept, frequently employed for ensuring balance in humanoid robots~\cite{5509348}, but has also been shown to hold true for human gait~\cite{popovic2005legged}. This approach ensures our generated motions are not only visually convincing but also more closely adhere to principles of biomechanics, making them suitable for a wide range of applications in realistic motion generation.

One of the key applications our model enables is retargeting of animation between characters with different identities. 
Existing methods typically generate human motions for a canonical body and then use a second character retargeting step to transfer the generated motions to the target body. 
Since classical retargeting methods rely on simple heuristics and ignore body shape, they tend to fail for extreme poses and complex motions involving significant body-on-ground and self-contact.
In contrast, \nameMethod effectively learns how people with different body shapes and proportions perform the same motion; see \cref{figure:tipman:teaser}.

Given an input motion for a particular identity, the \nameMethod encoder outputs an encoding in an identity-agnostic latent space. The decoder receives this encoding along with a target identity and outputs a motion that resembles the input motion but as performed by the new identity. We leverage solutions from unpaired image-to-image translation literature~\cite{zhu2017unpaired} to design a self-supervised loss that leverages cyclic consistency within the encoder-decoder step. The cycle-consistent formulation results in realistic motions given a target identity. 

We observe that the cycle-consistency alone is not enough for this task as the network may learn trivial solutions that ignore the target identity and output the same target motion as the source, while still satisfying the cycle consistency constraint. 
For example, merely copying the source motion to the target body will result in significant foot-sliding, ground penetration, floating, and dynamic instabilities.  
To prevent this, our key insight is to incorporate our IP and dynamic stability terms as training losses on the generated target body motions. This ensures the generated motions remain physically-plausible and dynamically stable, and encourages the network to use the conditioning body shape.
This makes \nameMethod the first data-driven human motion model that generates motions that are not only realistic but also physically plausible, dynamically stable, and tailored to the input body shape.

\section{Related Work}
\label{sec:relatedwork}
We categorize related work into several broad areas: previous works which transfer motion between different skeleton proportions or topologies, works which use physics simulation as a prior to constrain human motion generation, and works which synthesize or generate motion, often conditioned on various different desired parameters.

\vspace{0.25cm}

\noindent\textbf{Motion Transfer:}
Most industry methods of transferring motion between two characters assume that characters have either the same skeletal topology (but may differ in bone lengths) or a manual mapping between the two is provided. Motion is then largely transferred directly, without taking into account the further shape or identity difference between the characters. Simple heuristics such as inverse kinematics are used to remove any artifacts~\cite{motionbuilder,rokoko,803346}. There have been several attempts to apply human motion data to entirely different characters or creatures~\cite{https://doi.org/10.1111/cgf.12860,10.5555/1921427.1921453,DBLP:journals/corr/abs-2007-11341}, yet they tend to require paired data to function effectively which can be difficult to obtain at a large scale. Recent methods using Deep Learning have been developed which can transfer motion between different topologies~\cite{10.1145/3386569.3392462,10.1145/3610548.3618206} without paired data, or even between entirely new mesh shapes~\cite{https://doi.org/10.1111/cgf.12507}. Similarly, techniques have been developed which can re-target motion from other sources or data representations such as 2D videos~\cite{Aberman_2019}, or can take into account physical constraints such as floor contacts~\cite{9710501,10.1145/3359566.3360075}. However, these methods generally do not take into account the character's shape or identity beyond their skeleton proportions. While some attempts have been made to build retargeting systems which take into account character identities~\cite{10044431,10203899,gomes:hal-03257490,DBLP:journals/corr/abs-2003-07254}, these works are limited in scale, and only work on poses or very short windows of motion or on a small number of body types.

\noindent\textbf{Physics-based Motion Modeling:}
Physics-based motion generation, particularly through the use of reinforcement learning (RL) within physics engines, has emerged as a prominent method for creating physically plausible humanoid motions. This approach, leveraging RL, navigates the complex solution space of human motion, aiming to produce motions that adhere to physical laws. Common approaches in this direction include the development of locomotion skills and user-controllable policies for character animation through deep RL \cite{schulman2015high, schulman2017proximal, peng2017learning, bergamin2019drecon,peng2018deepmimic, peng2018sfv, won2020scalable, makoviychuk2021isaac}. Despite the principled framework RL offers, it comes with its limitations. The extensive training required due to the high-dimensional space of human motions, the reliance on reward functions over data for motion generation, and the computational expense of physics simulators present significant challenges. Also, these engines are typically non-differentiable black boxes, making them incompatible with data-driven learning frameworks~\cite{tripathi2023ipman,10.1145/3478513.3480527}. To overcome these challenges, physics-based trajectory optimization and motion imitation have been applied for 3D human pose estimation, further highlighting the significance of physics in capturing human dynamics \cite{yuan2020dlow, yuan2020residual, bergamin2019drecon, yuan2021simpoe, shimada2020physcap, shimada2021neural, SMPL:2015, zell2017joint, yi2022physical}. These simulators often utilize simplistic, non-differentiable models that fail to capture the intricacies of skin-level contact or muscle activations, leading to motions that, while physically plausible, lack the naturalness, diversity, and expressiveness found in data-driven approaches. We take inspiration from \cite{tripathi2023ipman} and biomechanical physics terms, and propose to combine the physical plausibility with our data-driven method to generate more accurate and lifelike human motions while considering body shape and physics.

\noindent\textbf{Data-Driven Human Motion Generation:} Early efforts in human motion generation utilized deterministic models, producing single motion outcomes and failing to capture the stochastic nature of human motions \cite{fragkiadaki2015recurrent, bergamin2019drecon, ghosh2017learning, gopalakrishnan2019neural, grenander1994representations, aksan2019structured, bao2022analytic, Guo_2022_CVPR, guo2020action2motion, holden2016motionsynthesis}. The shift towards deep generative models like GANs and VAEs marked a significant advancement, enabling the modeling of human motions' probabilistic nature conditioned on various inputs such as past motions \cite{barsoum2018hp, choi2021ilvr, aliakbarian2020stochastic, dhariwal2021diffusion,taheri2022goal, dockhorn2022genie, he2022nemf, Habibie2017ARV}, music or speech \cite{li2022danceformer, zhang2022motiondiffuse, li2021ai, dancing2music2019, Li2020LearningTG, ginosar2019gestures, bhattacharaya:2021}, and text or action labels \cite{ahuja2019language2pose, bhattacharya2021text2gestures, Guo_2022_CVPR, ren2022diffusion, guo2020action2motion, petrovich2021actor, Ahn2018Text2ActionGA, Ghosh_2021_ICCV}. The introduction of denoising diffusion models~\cite{bao2022estimating, tevet2023mdm, chen2023latent} represents a leap forward, merging the strengths of traditional generative models to achieve state-of-the-art performance in motion generation. Despite these advancements, a key challenge in shape conditioning remains: the lack of paired training data between body shape and motions. This leads most data-driven methods to forgo shape conditioning altogether. These approaches typically normalize training data, mainly from the AMASS~\cite{AMASS_2019} dataset, to a canonical skeleton or mean body shape ~\cite{chen2023latent, tevet2023mdm,  petrovich2021actor, petrovich2022temos, athanasiou2022teach, athanasiou2023sinc,rempe2021humor}, successfully learning the manifold of realistic motions for a canonical body, but often at the expense of physical and biomechanical realism. Such simplifications result in pronounced inconsistencies like foot sliding and ground penetration, undermining the realism necessary for applications in animation and virtual reality \cite{reitsma2003perceptual, hoyet2012push}.

\section{Method}
\label{sec:method}
Our goal is to learn an identity-conditioned motion model capable of generating 1) \emph{realistic}, 2) \emph{physically-plausible} and 3) \emph{dynamically-stable} human motions. Specifically, we represent a 3D human motion as a sequence of poses $\mathbf{P}_{1:T}=\mathbf{P}_1, \dots, \mathbf{P}_T$, where $T$ denotes the number of frames. We follow prior work~\cite{petrovich2021actor, petrovich2022temos, athanasiou2023sinc, athanasiou2022teach} and represent each pose $\mathbf{P}$ by the 3D SMPL~\cite{SMPL:2015} vertex mesh, $\mathbf{V}(\theta, \beta, \gender)$. 
We choose a mesh-based representation as any physical analysis of human motion requires accurate modeling of skin-level surface contact. SMPL is a convenient choice as it parameterizes the 3D mesh into disentangled pose, $\theta$, body shape, $\beta$ and gender, $\gender$ parameters, allowing explicit and independent control over the gender, body shape and pose. For ease of notation, we combine the body shape and gender into a single \emph{identity} parameter, $\identity = (\beta, \gender)$, and use it as the conditioning signal in our motion model. Given the target identity, $\identity_t$ and an arbitrary duration T, we generate a sequence consisting of the global root joint position, $\trans_t\in\mathbb{R}^3$ and the root-relative joint rotations, $\rot_t\in \mathbb{R}^{J\times6}$ in the 6D rotation representation~\cite{zhou20196d}, where $J=23$ is the number of SMPL joints and one global rotation. Although SMPL uses parent-relative rotations defined on the kinematic chain, we empirically observe that using root-relative joint rotations results in more stable gradients and faster convergence. Consequently, we convert the SMPL parent-relative pose parameters, to root-relative rotations to construct our motion features. Additionally, we process all sequences by removing the z-component of the first-frame root orientation $\rot^{z}_1$ and the horizontal root translation, $\trans^{x}_1$ and $\trans^{y}_1$. Doing this canonicalizes all sequences to start at the origin with the same forward facing direction and makes network training easier. Please refer to \supmat for a detailed description of our motion representation.    

\subsection{\nameMethod model architecture}

\begin{figure*}[ht]
\centering
\includegraphics[width=\linewidth]{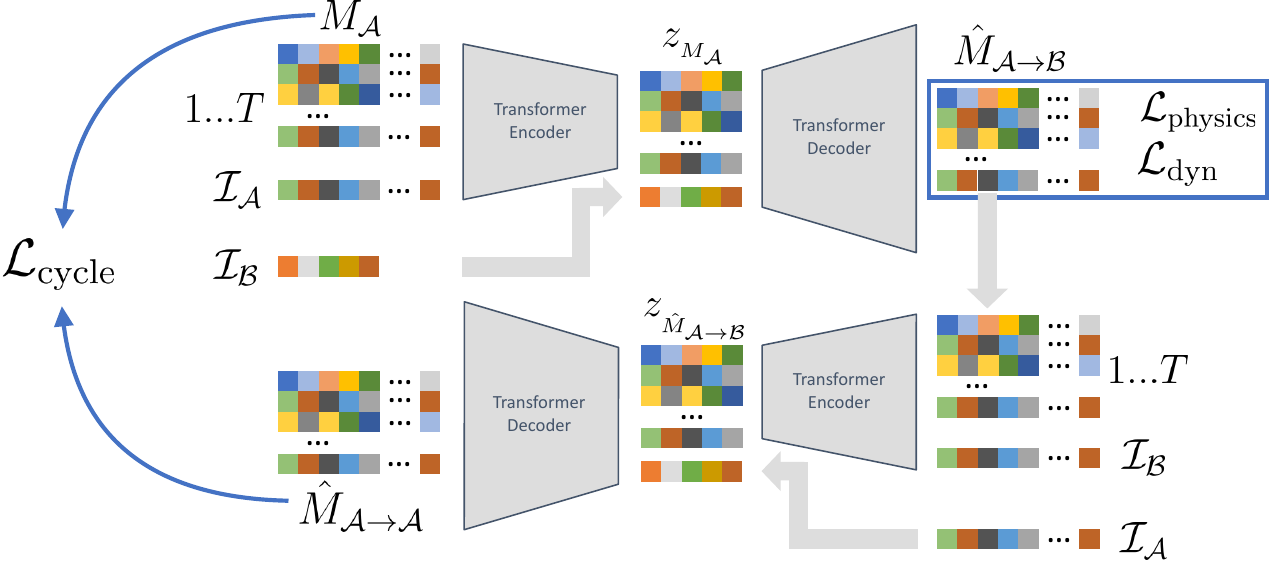}
\vspace{-1.5 em}
\caption{Visual representation of our architecture. The {\em Encoder} takes as input a motion $M_\idA$ and its associated identity $\identity_\idA$, and outputs a latent (identity invariant) encoding of the motion $z_{M_\idA}$. The {\em Decoder} takes as input the latent encoding of the motion $z_{M_\idA}$, along with a different identity $\identity_\idB$, and produces a retargeted motion appropriate for the given identity $\hat{M}_{\idA\to\idB}$. The same {\em Encoder} and {\em Decoder} are used with the original identity to produce a cycle loss $\mathcal{L}_{\text{cycle}}$, while a physics loss $\mathcal{L}_{\text{physics}}$ ensures the retargeted motion $\hat{M}_{\idA\to\idB}$ is realistic with respect to the given identity $\identity_\idB$ and prevents the cycle consistency loss from collapsing to a trivial solution. }
\label{figure:tipman:architecture}
\end{figure*}

\nameMethod is designed as a conditional Variation Auto-Encoder (c-VAE) network that generates sequential motion features in a non-autoregressive manner where we output motion features for $T$ consecutive frame in one-shot. Our choice of non-autoregressive training is driven by the observation that while auto-regressive approaches are effective in generating simple motions like walking, running, \etc, one-shot approaches yield better motion diversity~\cite{rempe2021humor}. Consequently, most text-to-motion approaches employ non-autoregressive generative modeling since they focus on generating diverse motions conditioned on text. Following this trend, we use a non-autoregressive training paradigm to output diverse motions conditioned on body shape. We use a Transformer~\cite{vaswani2017attention} architecture to obtain spatio-temporal embeddings from the input motion features. Next, we describe our motion encoder followed by the motion decoder (see \cref{figure:tipman:architecture} for an overview).    

\noindent\textbf{Motion encoder:} We extend the Transformer-based VAE motion encoder from TEMOS~\cite{petrovich2022temos} by conditioning it on identity features, $\identity$. Given a motion sequence $M_\idA$ of arbitrary length, T, and the conditioning signal, $\identity_\idA$ of the source identity, our encoder $\mathcal{E}$ outputs distribution tokens $\mu$ and $\Sigma$ for the shape-conditioned motion latent space. Using the reparameterization trick~\cite{kingma2014vae}, we sample the latent vector $z_{M_\idA}\in R^d$, embedding the input features into a $d$-dimensional latent space. To represent the temporal ordering in the input sequence, we use positional encodings in the form of sinusoidal functions and concatenate them with the input features~\cite{vaswani2017attention}.        

\noindent\textbf{Motion decoder:} The decoder takes as input the latent code $z_{M_\idA}$ and the target identity,  $\identity_\idB$ and generates a sequence of 3D poses $\mathbf{\hat{P}}_{1:T}$ representing the source 3D motion as performed by an individual with the target body shape and gender. 
Our motion decoder $\mathcal{D}$ is built on a Transformer architecture, which incorporates time information through $T$ sinusoidal positional embeddings as queries. A concatenated combination of latent vectors and identity features serves as the key-value pair. Our decoder architecture mirrors the motion encoder, except for the input and output layers. 

\subsection{Self-supervised shape-conditioned training}

Training a shape-conditioned c-VAE in a fully supervised manner requires paired identity and motion data. While datasets like AMASS~\cite{AMASS_2019} contain a large collection of 3D motion capture sequences, often containing diverse motions performed by the same person, we seldom find different identities performing the same motion. Such pairs are essential for any motion model to effectively disambiguate motion from body type. This lack of \emph{paired} data impedes training a shape-conditioned model. In fact, most existing methods capable of generating human motions ignore the body shape, generating motions only for the canonical skeleton and mean body shape using SMPL's \emph{neutral} gender body model.    

To circumvent the lack of paired data in AMASS, we draw inspiration from the success in image-to-image translation~\cite{zhu2017unpaired} and use a novel self-supervised training strategy that leverages cycle-consistency in the shape-motion space. As shown in \cref{figure:tipman:architecture}, we randomly sample a motion capture sequence for identity $\idA$ and extract motion features $M_\idA$ comprising the global root-joint translation $\trans_{\idA}$ and the root-relative joint rotations $\rot_{\idA}$. We project the identity features, $\identity_\idA=({\beta_\idA, \gender_\idA})$ using a linear layer. The identity features are concatenated with the motion features and fed to the motion encoder which embeds them in a shape-agnostic latent code, $z_{M_\idA}$. The \nameMethod decoder takes the latent code $z_{M_\idA}=\mathcal{E}(M_\idA, \identity_\idA)$ as input and a randomly sampled target identity, $\idB$, with the projected $\identity_\idB$ as the new conditioning input to the decoder. The task of the decoder is to generate the root-joint translations $\hat{\trans}_{\idA\to\idB}$ and the joint rotations $\hat{\rot}_{\idA\to\idB}$ representing $\hat{M}_{\idA\to\idB}=\mathcal{D}(z_{M_\idA}, \identity_\idB)$, \ie \emph{the motion $M_\idA$ in the style of the identity of $\idB$}. 

Since we lack any explicit ground truth to supervise $\hat{M}_{\idA\to\idB}$, as illustrated in \cref{figure:tipman:architecture}, we employ cycle-consistency by reversing the forward step, this time using $\hat{M}_{\idA\to\idB}$ as the source motion and $\idA$ as the target identity. We input $\hat{M}_{\idA\to\idB}$ and $\identity_\idB$ to the \nameMethod encoder and extract latent code, $z_{\hat{M}_{\idA\to\idB}}$. The latent code, along with projected $\identity_\idA$ are fed to the decoder resulting in $\hat{M}_{\idA\to\idA}=\mathcal{D}(z_{\hat{M}_{\idA\to\idB}}, \identity_\idA)$. In the full cycle, since the motion style remains the same even as identities change, the same identity features should result in the same motion. Consequently, the reconstructed $\hat{M}_{\idA\to\idA}$ should match the source motion $M_{\idA}$ and we define the cycle-consistency loss as $\mathcal{L}_{\text{cycle}} = \mathcal{L}_{\text{rot}} + \mathcal{L}_{\text{pos}}$ where $\mathcal{L}_{\text{rot}}$ computes the geodesic distance in the rotational space by converting 6D joint rotations, $\rot_{t_\idA}$ and $\hat{\rot}_{t_{\idA\to\idA}}$ to rotation matrices $R_{t_\idA}$ and $\hat{R}_{t_{\idA\to\idA}}$ using the Gram-Schmidt process~\cite{zhou20196d}. $\mathcal{L}_{\text{pos}}$ is the smooth L1 loss between source and reconstructed root joint positions, $\trans_{t_{\idA}}$ and $\hat{\trans}_{t_{\idA\to\idA}}$, Specifically, 
\begin{equation}
\mathcal{L}_{\text{cycle}} = \mathcal{L}_{\text{rot}} + \mathcal{L}_{\text{pos}} \\
\end{equation}
\begin{equation}
\mathcal{L}_{\text{rot}} = \sum_{t=1}^T\arccos\frac{\Tr\Big(R_{t_\idA}(\hat{R}_{t_{\idA\to\idA}})^{-1}\Big) - 1}{2},\quad \mathcal{L}_{\text{pos}} = \sum_{t=1}^T\|\trans_{t_{\idA}} - \hat{\trans}_{t_{\idA\to\idA}}\|_1.
\end{equation}

\subsection{Intuitive-physics (IP) terms}

Our motion encoder aggregates spatio-temporal features over successive frames to learn a shape-agnostic latent embedding by disentangling the motion ``style'' from identity-specific attributes. The decoder, in turn, leverages the shape-agnostic latent code and maps the motion style to a new target body. While intuitive, $\mathcal{L}_\text{cycle}$, however, is not enough as training with only $\mathcal{L}_{\text{cycle}}$ is prone to trivial solutions. Without special care, the encoder-decoder architecture could learn to generate identical motion $M_{\idA}\approx \hat{M}_{\idA\to\idB}\approx \hat{M}_{\idA\to\idA}$ at intermediate steps, ignoring the identity conditioning while na\"ively minimizing $\mathcal{L}_{\text{cycle}}$. To alleviate this, as shown in \cref{figure:tipman:architecture}, we incorporate intuitive physics terms $\mathcal{L}_{\text{physics}}$ on $\hat{M}_{\idA\to\idB}$ that address physical inconsistencies such as ground penetration, floating meshes and foot sliding. If \nameMethod na\"ively copies the same source motion $M_\idA$ on the target body $\idB$, it would result in motions $\hat{M}_{\idA\to\idB}$ that have significant ground penetration, floating meshes and foot sliding.

Our intuitive physics terms are fast, simple and fully-differentiable. Following \cite{yuan2023physdiff}, we design IP terms to address penetration, float, and foot sliding individually. Our $\mathcal{L}_{\text{penetrate}}$ minimizes the per-frame distance of the lowest vertex \emph{below} the ground from the ground plane. $\mathcal{L}_{\text{float}}$ minimizes the per-frame distance of the lowest vertex \emph{above} the ground from the ground plane. The foot sliding loss, $\mathcal{L}_{\text{slide}}$, minimizes the horizontal x-y component of the foot joint velocities if they are determined to be in ground contact using a distance threshold from the ground. We collate them together as $\mathcal{L}_{\text{physics}} = \mathcal{L}_{\text{penetrate}} + \mathcal{L}_{\text{float}} + \mathcal{L}_{\text{slide}}$.       

\subsection{Dynamic stability term}

In the real world, motion is the result of internal muscular forces and external forces acting on the body and the surrounding scene. Human bodies are typically \emph{stable}, i.e. they have the ability to control their body position and momentum during movement without falling over.

Tripathi~\etal~\cite{tripathi2023ipman} successfully use the notion of static stability in 3D human pose and shape estimation to output physically-plausible and biomechanically stable poses from RGB images. In static poses, a body is considered stable if the gravity-projection of the center of mass (CoM) onto the ground is within the base of support (BoS). The base of support is defined as the convex hull of all points in contact with the ground. Since the base of support requires a convex hull computation that is not easily differentiable,~\cite{tripathi2023ipman} minimize the distance between an estimated center of pressure (CoP) and the projected CoM instead, and use it as a proxy static stability loss or energy term which is minimized during training and optimization. However, this \emph{static} treatment of stability is only applicable to static poses. Humans are highly dynamic by nature and we need a general treatment of stability analysis that extends to all scenarios involving human movement and locomotion.

Dynamic stability extends this concept to bodies in motion. We follow the concept of zero-moment point (ZMP)~\cite{vukobratovic2004zmp}, which has been widely used in robotics and biomechanics~\cite{kondak2003control,popovic2005legged}.
Assuming flat ground, the ZMP is the point on the ground where the horizontal component of the moment of ground reaction force is zero. If this point lies within the base of support, the ZMP is equivalent to the center of pressure and the motion is considered dynamically stable (see \supmat video for an example in human gait).

The ZMP is defined as a function of the CoM's acceleration and the net moment torques along the CoM and can be computed in closed form in a fully differentiable manner:
\begin{equation}
\mathcal{Z} = \mathcal{C}_m - \frac{n \times \mathcal{M}_{\mathcal{C}_m}^{gi}}{\mathcal{F}^{gi} \cdot n }
\end{equation}
where $\mathcal{C}_m$ is the projection of the center of mass onto the ground plane, and $n$ is the normal to the ground plane. $\mathcal{F}^{gi}$ is force of inertia calculated as
\begin{equation}
\mathcal{F}^{gi} = mg - ma_\mathcal{G}
\end{equation}
with $m$ being the total mass of the body, $g$ the acceleration of gravity, and $a_\mathcal{G}$ the acceleration of the center of mass $\mathcal{G}$. $\mathcal{M}_{\mathcal{C}_m}^{gi}$ is the moment around the projected center of mass $\mathcal{C}_m$
\begin{equation}
\mathcal{M}_{\mathcal{C}_m}^{gi} = \overrightarrow{\mathcal{C}_m\mathcal{G}} \times mg - \overrightarrow{\mathcal{C}_m\mathcal{G}} \times ma_\mathcal{G} - \dot{\mathcal{H}}_\mathcal{G}
\end{equation}

where $\dot{\mathcal{H}}_\mathcal{G}$ is the rate of change of angular momentum or torque at the center of mass.

We calculate the total body mass $m$ by using the volume of the SMPL mesh as a proxy for total weight. To calculate center of mass $\mathcal{G}$ and its acceleration as well as the moment $\mathcal{M}_{\mathcal{C}_m}^{gi}$, we distribute the total mass m to point masses at the vertices of the body mesh proportional to the volume of the body part they are part of. The accelerations are calculated numerically using the central differences. Please refer to \supmat for detailed derivations of the formulas.

Similar to Tripathi~\etal~\cite{tripathi2023ipman} we use an estimation of the center of pressure as proxy for calculating the distance to the base of support. The center of pressure $\mathcal{C}_P$ is calculated as weighted average of all vertices close to ground plane in a frame.

For dynamically stable motions, the ZMP and CoP should coincide. We, therefore, minimize the distance between ZMP and CoP and define our dynamic stability loss as
\begin{equation}
\mathcal{L}_{\text{dyn}} = \rho(\|\mathcal{C}_P - \mathcal{Z}\|_2)
\end{equation}
where $\rho$ is the Geman-McClure penalty function~\cite{geman1987statistical} which stabilizes training by making $\mathcal{L}_{\text{dyn}}$ robust to noisy ZMP estimates. 

\camready{Dynamic stability computation requires ground support. For sequences where the human is not supported by the ground, \ie lowest vertex $>$25 cm, we disable the dynamic stability term during training. Thus, although the dynamic stability term is designed to help grounded sequences, it does not hurt non-grounded ones.}

\subsection{Latent embedding losses}

To enable motion generation at inference time, we regularize the distributions of the latent embedding spaces, $z_{M_\idA}=\mathcal{N}(\mu_\idA, \Sigma_\idA)$ and $z_{M_{\idA\to\idB}}=\mathcal{N}(\mu_{\idA\to\idB}, \Sigma_{\idA\to\idB})$ to be similar to the normal distribution $\psi=\mathcal{N}(0, I)$ by minimizing the Kullback-Leibler (KL) divergence via
\begin{equation}
\mathcal{L}_{\text{KL}} = \text{KL}(z_{M_\idA},\psi) + \text{KL}(z_{M_{\idA\to\idB}},\psi)  \\
\end{equation}

Since the \nameMethod latent embeddings encode motion style rather than identity-specific attributes, we also encourage the embeddings $z_{M_\idA}\sim\mathcal{N}(\mu_\idA, \Sigma_\idA)$ and $z_{M_{\idA\to\idB}}\sim\mathcal{N}(\mu_{\idA\to\idB}, \Sigma_{\idA\to\idB})$ to be as close as possible to each other via the the following cycle-consistent L1 loss:
\begin{equation}
\mathcal{L}_{\text{E}} = \|z_{M_\idA} - z_{M_{\idA\to\idB}}\|_1
\end{equation}

The resulting total loss in \nameMethod training is the weighted sum of all the individual loss terms:
\begin{equation}
\mathcal{L} = \lambda_{\text{cycle}}\ \mathcal{L}_{\text{cycle}}+\lambda_{\text{physics}}\ \mathcal{L}_{\text{physics}} + \lambda_{\text{dyn}}\ \mathcal{L}_{\text{dyn}} + \lambda_{\text{KL}}\ \mathcal{L}_{\text{KL}} + \lambda_{\text{E}}\ \mathcal{L}_{\text{E}}
\end{equation}

The loss weights are determined empirically and set to $\lambda_{\text{cycle}}=1$, $\lambda_{\text{physics}}=1$, $\lambda_{\text{dyn}}=0.0001$, $\lambda_{\text{KL}}=10^{-5}$ and $\lambda_{\text{E}}=10^{-2}$.

\section{Experiments}
\label{sec:experiments}
We first discuss our data processing and implementation details in (\cref{sec:tipman:impl_details}). Next, we introduce baselines and the evaluation metrics (\cref{sec:tipman:baselines}) used in our comparisons. Then, we discuss quantitative, perceptual and qualitative comparisons of our method with baselines (\cref{sec:tipman:comparisons}) and present an ablation study (\cref{sec:tipman:ablation}).

\subsection{Data and implementation details}
\label{sec:tipman:impl_details}

For training, we use the AMASS dataset which contains 480 unique gender identities out of which 274 are male and 206 are female with diverse body shapes and sizes. Please refer to \supmat for a full analysis on the diversity and distribution of body shape $\beta$ parameters in AMASS. We first subsample raw \smpl-H sequences from AMASS~\cite{AMASS_2019} to 20 fps following Guo~\etal~\cite{Guo_2022_CVPR}. We augment the data by mirroring sequences left-to-right. We exclude sequences where the feet are more than 20 cm above the ground and where the lowest vertex across all frames is not grounded. This is to ensure ground support as it is an essential component for dynamic stability. We observed that normalizing sequences for consistent facing and start position in the first frame helped with training. 
We then extract input features by converting the root orientation and joint rotations to 6D form~\cite{zhou20196d} and concatenate root translation, betas and gender. 
In addition, we augment the AMASS dataset by applying left-right flip augmentation to the pose parameters, effectively doubling the amount of training data. We show a step-by-step visualization of our data-processing pipeline in \supmat\footnote{All datasets were obtained and used only by the authors affiliated with academic institutions.} 

We train our models for 1300 epochs with the AdamW~\cite{kingma2015adam, Loshchilov2017DecoupledWD} optimizer using a fixed learning rate $10^{-5}$ and a batch size of $60$. Both our encoder and decoder consist of 6 transformer layers. We train with sequence length T = 200 frames on arbitrary length clips sampled from AMASS. For clips longer or shorter than T frames, we extend or clip the sampling interval by either including neighboring frames or dropping boundary frames. For short AMASS videos with $<$ T frames, we repeat the last frame. Please refer to \supmat for more details.     

\subsection{Baselines and evaluation metrics}
\label{sec:tipman:baselines}

We focus on the task of shape-conditioned motion reconstruction for evaluating the performance of \nameMethod. Since no existing baseline directly addresses shape-conditioned motion reconstruction, we create new baselines by combining a state-of-the-art motion generation model, TEMOS~\cite{petrovich2022temos} and retargeting its output motions to a target body shape using 1) na\"ive retargeting and 2) using the commercial retargeting system, Rokoko~\cite{rokoko}. For these experiments, we reconstruct the same motions from the AMASS test-split for both the TEMOS baselines and \nameMethod. 

TEMOS generates motions for a canonicalized mean-shape SMPL body by directly regressing the pose and global root-joint translation. For a fair comparison, we use its ``unconditional'' variant which does not require any text inputs. For obtaining motions for a target body, we do simple retargeting where we randomly sample identities from AMASS and na\"ively \emph{copy} the target $\beta_\idB$ and gender $\gender_\idB$ parameters to the motions obtained from TEMOS. We call this baseline, ``TEMOS-Simple''. Intuitively, na\"ively copying a new identity to a neutral mean-shape body motion will result in ground peneration, floating and foot sliding artefacts. We address ground penetration by translating the whole motion sequence such that the lowest vertex in the sequence is on the ground. We refer to this baseline as ``TEMOS-Simple-G''. For another strong baseline, we use the Rokoko retargeting system to retarget TEMOS generated motions to the target body. We call this baseline ``TEMOS-Rokoko'', with ``TEMOS-Rokoko-G'' being the variant where we ground the Rokoko output sequences as described above. For consistency in evaluation, given an input motion, we sample the same target identities across our method and all baselines. 

\noindent\textbf{Evaluation Metrics.} For evaluating the physical plausibility of generated motions, we use the physics-based metrics suggested by Yuan~\etal~\cite{yuan2023physdiff}. The \emph{Penetrate} metric measures ground penetration by computing the distance (in cm) between the ground and the lowest body mesh vertex \emph{below} the ground. \emph{Float} measuring the amount of unsupported floating by computing the distance  (in cm) between the ground and the lowest body mesh vertex \emph{above} the ground. \emph{Skate} measures foot skating by computing the percentage of adjacent frames where the foot joints in contact with the ground have a non-zero average velocity. We also report two metrics for measuring dynamic stability. \emph{Dyn. Stability} computes the percentage of frames where the ZMP is outside the base of support. The \emph{BoSDist} metric measures the distance of the ZMP to the closest edge of the BoS convex hull, if the ZMP lies outside the BoS, indicating the pose is dynamically unstable. For details, please refer to \supmat   

\subsection{Comparison to baselines}
\label{sec:tipman:comparisons}

\begin{figure*}[ht]
\centering
\includegraphics[width=\linewidth]{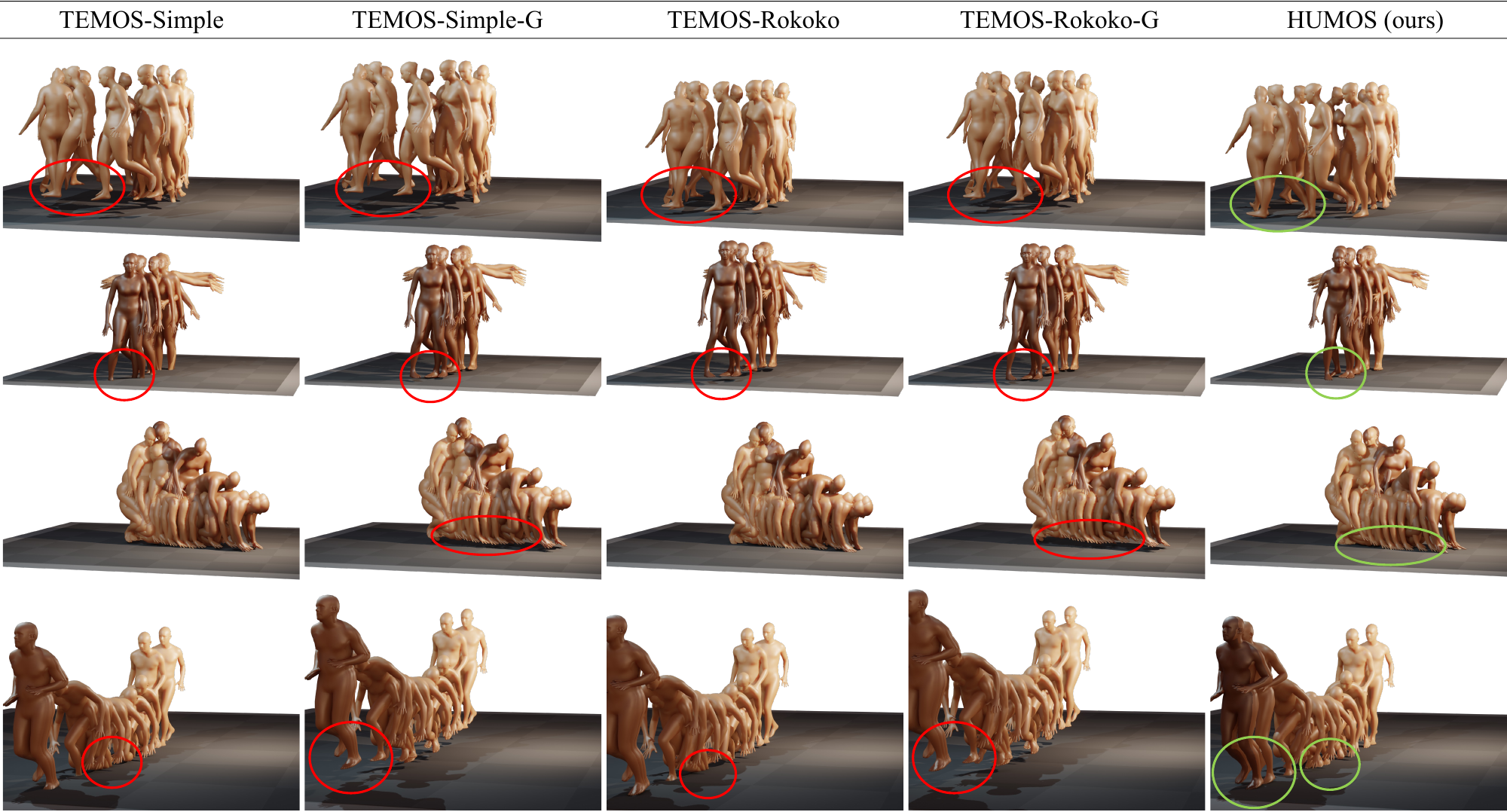}
\caption{Qualitative comparison of shape-conditioned motion generation. Each row represents generations across different methods for a unique body shape and gender. \nameMethod generated motions are more realistic, physically plausible and dynamically stable compared to baselines. The red circles on the baseline methods highlight issues such as floating, penetrations, and foot skating, compared to more realistic results on highlighted in green with \nameMethod. \faSearch~\textbf{Zoom in}.} 
\label{figure:tipman:qualitative_comparison}
\end{figure*}

\noindent\textbf{Quantitative.} We summarize our main results in \cref{tab:sota_comparison}. As we lack ground-truth motions for the target body shape, we rely on physics and stability metrics to compare our method with baselines. Our method substantially outperforms on all metrics except on \emph{Penetrate}. As expected, the lowest ground penetration is observed for TEMOS-Simple-G and TEMOS-Rokoko-G as both were specifically altered to ground the sequence. However, this comes at the cost of increasing the \emph{Float} metric. In contrast, \nameMethod simultaneously improves both \emph{Penetrate} and \emph{Float} indicating that the network learns to modify body pose (\eg foot tilt) in addition to learning the correct global translation for grounding the motion. \nameMethod also improves over baselines on foot skating, achieving a $\sim7.3\%$ \emph{Skate} compared to $20\%$ and $27\%$ for the TEMOS-Rokoko and TEMOS-Simple baselines. \nameMethod's motions are also dynamically stable in $71.9\%$ of all frames, a significant improvement of $16\%$ over the closest baseline. Even for dynamically unstable poses, when the ZMP is outside the BoS, it is close to the BoS edge as indicated by the low \emph{BoSDict} metric for our method.

\noindent\textbf{Qualitative.} We provide additional qualitative comparisons with baselines in \cref{figure:tipman:qualitative_comparison}. Each row represents the same pair of source motion and target body across all methods. We highlight physical plausibility issues such as foot-skate, ground penetration and floating in \textcolor{red}{red}. The \textcolor{green}{green} highlighted region points to the improvement in \nameMethod's results over baselines. \nameMethod motions showcase realistic ground support. To better evaluate our performance on shape-conditioned motion generation, foot-sliding and dynamic stability, please watch our \supmat video. 

\begin{table*}[ht]
\centering
\aboverulesep=0ex
\belowrulesep=0ex
\caption{Comparison of \nameMethod with baselines on the shape-conditioned motion reconstruction task.}
\label{tab:sota_comparison}
\begin{adjustbox}{width=1\textwidth}
\setlength{\tabcolsep}{4pt}
\begin{tabular}{@{}l|c|c|c|c|c@{}}
\toprule
\rule{0pt}{1.1EM}
       \textbf{Method} & 
       \multicolumn{1}{c|}{\textbf{Penetrate (cm)} $\downarrow$}                                & 
       \multicolumn{1}{c|}{\textbf{Float (cm)} $\downarrow$}                                    &
       \multicolumn{1}{c|}{\textbf{Skate (\%)} $\downarrow$}                                    &
       \multicolumn{1}{c|}{\textbf{Dyn. Stability (\%)}$\uparrow$}   &                 
       \multicolumn{1}{c}{\textbf{BoS Dist (cm)} $\downarrow$} \\
\midrule
\rule{0pt}{1.1EM}
TEMOS~\cite{petrovich2022temos}-Simple  &  6.82 & 6.55 & 27.07 & 45.85 & 16.94   \\ %
\ TEMOS~\cite{petrovich2022temos}-Simple-G & \textbf{0.75} & 4.39 & 27.07 & 45.85 & 16.94   \\ %
\ TEMOS~\cite{petrovich2022temos}-Rokoko~\cite{rokoko_studio_live_blender}  & 4.14 & 3.85 & 20.05 & 55.92 & 16.58   \\ %
\ TEMOS~\cite{petrovich2022temos}-Rokoko~\cite{rokoko_studio_live_blender}-G  & \textbf{0.75} & 4.44 & 20.05 & 55.92 & 16.58 \\
\midrule
\ \nameMethod  & 1.23 & \textbf{1.04} & \textbf{7.37}  & \textbf{71.9} & \textbf{14.62} \\
\hline
\end{tabular}
\end{adjustbox}
\vspace{-1em}
\end{table*}

\noindent\textbf{Perceptual Study.} To evaluate realism of the generated motions given the target body shape, we perform a human study on Amazon Mechanical Turk (AMT). We randomly select 30 videos generated from our method and the next closest baselines, ``TEMOS-Rokoko'' and ``TEMOS-Rokoko-G''. The participants are shown a single video and after watching the whole video at least once, are allowed to select their response to the question ``How realistic is the motion given this body shape?'' on a Likert scale of scores between 5 (completely realistic) to 1 (completely unrealistic). Each rating task was completed by 25 participants. In the study, we added 4 catch trials, 2 containing ground-truth AMASS motions and 2 containing significant ground penetration. 17 out of 75 participants who failed the catch trials were excluded from our study. As shown in \cref{tab:tipman:perceptual} participants prefer \nameMethod motions and give it an average rating of $3.64$ out of $5$, compared to $3.25$ for TEMOS-Rokoko and $3.19$ for TEMOS-Rokoko-G. Curiously, between the two baselines, participants preferred the motions without a heuristics-based grounding indicating that the use of heuristics for motion retargeting struggles with generalization. We include more details about the study and layout in \supmat 

\begin{table}[h!]
\centering
\aboverulesep=0ex
\belowrulesep=0ex
\vspace{-0.2em}
    \caption{Perceptual study comparing \nameMethod with two closest baselines. Given a video of a generated motion, participants select 5-point 
    ratings for the question ``How realistic is the motion given this body shape?''}
    \label{tab:tipman:perceptual}
    \begin{tabular}{l|c|c}
    \toprule
      \textbf{Method} & \textbf{Average Rating $\uparrow$} & \textbf{Std. Dev. $\downarrow$} \\
      \hline
      TEMOS-Rokoko & 3.25 & 1.26\\
      TEMOS-Rokoko-G & 3.19 & 1.27\\
      \hline
      \nameMethod & \textbf{3.64} & \textbf{1.11} \\
    \end{tabular}
\end{table}

\subsection{Ablation Study}
\label{sec:tipman:ablation}

We evaluate the importance of our key contributions, $\mathcal{L}_\text{cycle}$, $\mathcal{L}_\text{physics}$ and $\mathcal{L}_\text{dyn}$ in \cref{tab:ablation}. As shown, $\mathcal{L}_\text{cycle}$ alone achieves a significant $\sim33\%$ improvement in \emph{Penetrate}, $\sim32\%$ in \emph{Float}, $\sim25\%$ in \emph{Skate}  over the TEMOS-Rokoko baseline indicating that our cycle-consistent training paradigm is effective in training \nameMethod for the shape-conditioned motion generation task. Adding $\mathcal{L}_\text{physics}$ further improves physical plausibility, resulting in the biggest improvement in foot skating ($\sim47\%$). While both $\mathcal{L}_\text{cycle}$ and $\mathcal{L}_\text{physics}$ help, adding $\mathcal{L}_\text{dyn}$ results in the best \nameMethod configuration across all metrics. With all losses active, \nameMethod motions are dynamically stable $71.9\%$ the times.

\begin{table*}[ht]
\centering
\aboverulesep=0ex
\belowrulesep=0ex
\caption{Ablation study comparing the improvements from cycle-consistent training ($\mathcal{L}_{\text{cycle}}$), physics losses ($\mathcal{L}_{\text{physics}}$) and the dynamic stability term ($\mathcal{L}_{\text{dyn}}$).}
\label{tab:ablation}
\begin{adjustbox}{width=1\textwidth}
\setlength{\tabcolsep}{4pt}
\begin{tabular}{@{}l|c|c|c|c|c@{}}
\toprule
\rule{0pt}{1.1EM}
       \textbf{Method} & 
       \multicolumn{1}{c|}{\textbf{Penetrate (cm)} $\downarrow$}                                & 
       \multicolumn{1}{c|}{\textbf{Float (cm)} $\downarrow$}                                    &
       \multicolumn{1}{c|}{\textbf{Skate (\%)} $\downarrow$}                                    &
       \multicolumn{1}{c|}{\textbf{Dyn. Stability (\%)}$\uparrow$}   &                 
       \multicolumn{1}{c}{\textbf{BoS Dist (cm)} $\downarrow$} \\
\midrule
\rule{0pt}{1.1EM}
TEMOS-Rokoko  & 4.14 & 3.85 & 20.05 & 55.92 & 16.58   \\ %
\midrule
\ $\mathcal{L}_{\text{cycle}}$  &  2.74 & 2.62 & 15.04 & 64.00 & 16.96   \\ %
\ $\mathcal{L}_{\text{cycle}}$ + $\mathcal{L}_{\text{physics}}$ & 1.55 & 1.44 & 7.93 & 67.82 & 16.41   \\ %
\ $\mathcal{L}_{\text{cycle}}$ + $\mathcal{L}_{\text{physics}}$ + $\mathcal{L}_{\text{dyn}}$  & \textbf{1.23} & \textbf{1.04} & \textbf{7.37}  & \textbf{71.9} & \textbf{14.62} \\ %
\hline
\end{tabular}
\end{adjustbox}
\vspace{-2em}
\end{table*}

\section{Conclusion}
\label{sec:conclusion}
In this paper we presented a method for shape-conditioned motion generation that used a set of physically inspired constraints to allow for self-supervised disentanglement of character motion and identity. This allows for motion generation and retargetting of a higher quality than previous methods both qualitatively and quantitatively.

In terms of limitations, although our method represents an improvement over previous work there are still motion artefacts introduced by the model. Additionally, the differences in the style of motion produced by characters of very different body shapes remain subtle. This may be due to the limited shape diversity in the training set. In the future it would be interesting to examine how this data distribution affects the diversity and generalization capabilities of the model. We also do not take into account self-penetrations that may arise during shape-conditioned motion generation. Addressing this would be another promising direction for future work. 
\camready{While human motion is influenced by both body shape and individual motion style, we only consider body shape. Motion style includes factors like emotional state, physiological impediments, societal influences, and personal biases, which are not annotated in existing mocap datasets. With style-specific annotations, it would be useful to extend \nameMethod to include style attributes as additional conditioning signals.}

\label{sec:acknowledgements}
\camready{\textbf{Acknowledgements.} We sincerely thank Tsvetelina Alexiadis, Alpar Cseke, Tomasz Niewiadomski, and Taylor McConnell for facilitating the perceptual study, and Giorgio Becherini for his help with the Rokoko baseline. We are grateful to Iain Matthews, Brian Karis, Nikos Athanasiou, Markos Diomataris, and Mathis Petrovich for valuable discussions and advice. Their invaluable contributions enriched this research significantly.}

\bibliographystyle{config/splncs04}
\bibliography{config/BIB}

\clearpage
\appendix
{\noindent\LARGE\textbf{Appendix}}
\newline
\newline
\renewcommand{\thefigure}{S.\arabic{figure}}
\renewcommand{\thetable}{S.\arabic{table}}
\renewcommand{\theequation}{S.\arabic{equation}}
\setcounter{figure}{0}
\setcounter{table}{0}
\setcounter{equation}{0}
In the supplementary materials, we provide a detailed derivation of the ZMP and the dynamic stability term (\cref{supmat:deriv}), analyze the effect of body shape on motion (\cref{supmat:body_shape_effect}), provide additional qualitative results (\cref{supmat:additional_qualitative}), ablations for the latent embedding losses (\cref{supmat:latent_ablation}), a discussion on AMASS shape diversity (\cref{supmat:amass}), and finally additional implementation details (\cref{supmat:additional_implementation_details}).  
\\

\noindent \textbf{Video.} Our research focuses on humans in motion with diverse body shapes and sizes, making motion a critical aspect of our results. Given the difficulty of conveying motion quality through a static document, we strongly recommend that readers view the provided supplemental video for an in-depth overview of our methodology and findings.

\section{Detailed derivation for the Zero Moment Point (ZMP) and the dynamic stability term}
\label{supmat:deriv}

Before we compute the ZMP, we first compute the body Center of Mass (CoM) by adapting the CoM formulation of Tripathi~\etal~\cite{tripathi2023ipman} to dynamic humans. For every sequence, we use their body part segmentation and the differentiable \emph{``close-translate-fill''}~\cite{tripathi2023ipman} to compute per-part volumes $\volumeParti$ by splitting the mesh in the first frame into 10 parts. Using the per-part volumes, the CoM is calculated for time instance $t$, as a volume weighted-average of $N_U=6890$ mesh vertex points.  
\begin{equation}    %
    \mathcal{G}_t = \frac{\sum_{i=1}^{N_U} \volumePartvi v_{i_t}}{\sum_{i=1}^{N_U} \volumePartvi}    \text{,}
\end{equation}%
The acceleration of the CoM, $a_\mathcal{G}$, is obtained using the central difference as,   
\begin{equation}
    a_{\mathcal{G}_t} = \frac{\mathcal{G}_{t+1} - 2\mathcal{G}_t + \mathcal{G}_{t-1}}{\Delta t^2}
\end{equation}
With $a_{\mathcal{G}_t}$, the force of inertia, $\mathcal{F}^{gi}$, is computed as
\begin{equation}
\mathcal{F}^{gi} = mg - ma_\mathcal{G}
\end{equation}
where $m$ is the body mass. 
The moment around the projected CoM, $\mathcal{C}_m$, is 
\begin{equation}
\mathcal{M}_{\mathcal{C}}^{gi} = \overrightarrow{\mathcal{C}_m\mathcal{G}} \times mg - \overrightarrow{\mathcal{C}_m\mathcal{G}} \times ma_\mathcal{G} - \dot{\mathcal{H}}_\mathcal{G}
\end{equation}
where $\overrightarrow{\mathcal{C}_m\mathcal{G}}$ is the vector joining the projected CoM, $\mathcal{C}_m$ with the actual CoM, $\mathcal{G}$ and $\dot{\mathcal{H}}_\mathcal{G}$ is the rate of change of angular momentum at the CoM. For $\dot{\mathcal{H}}_\mathcal{G}$, we equally distribute the total $m$ to point masses at the vertices of the body mesh proportional to the volume of the body part they are part of. The per-vertex mass and acceleration is 
\begin{equation}
m_{v_i} = \frac{\volumePartvi}{\sum_{i=1}^{N_U} \volumePartvi}m, \quad
a_{v_i} = \frac{v_{i_{t+1}} - 2v_{i_t} + v_{i_{t-1}}}{\Delta t^2}
\end{equation}
And $\dot{\mathcal{H}}_\mathcal{G}$ is
\begin{equation}
\dot{\mathcal{H}}_\mathcal{G} = \sum_{i=1}^{N_U}\overrightarrow{v_i\mathcal{G}} \times m_{v_i}a_{v_i}
\end{equation}
 Finally, the ZMP is computed in closed-form as 
\begin{equation}
\mathcal{Z} = \mathcal{C}_m - \frac{n \times \mathcal{M}_{\mathcal{C}_m}^{gi}}{\mathcal{F}^{gi} \cdot n }
\end{equation}
For CoP computation, we follow Tripathi~\etal and uniformly sample the body mesh into $N_p=20000$ uniformly sampled surface points. We, then, use their heuristic pressure field to compute per-point, $p_i$, pressure as 
\begin{eqnarray}
    \rho_i &=& 
    \begin{cases}
        1 - \alpha h(p_i) & \text{if~} h(p_i) <    0  \text{,} \\
        e^{-\gamma  h(p_i)} & \text{if~}  h(p_i) \geq 0  \text{,}
    \end{cases} 
\end{eqnarray}
where $\alpha=100$ and $\gamma=10$ are scalar hyperparameters set empirically. The CoP is computed as,
\begin{eqnarray}
\mathcal{C}_p &=& \frac{\sum_{i=1}^{N_p}\rho_i p_i}{\sum_{i=1}^{N_p}\rho_i} \text{.}
\end{eqnarray}
With the ZMP and the CoP, known the dynamic stability loss is defined as, 
\begin{equation}
\mathcal{L}_{\text{dyn}} = \rho(\|\mathcal{C}_P - \mathcal{Z}\|_2)
\end{equation}
where $\rho$ is the Geman-McClure penalty function~\cite{geman1987statistical}.

\section{Effect of body shape on motion}
\label{supmat:body_shape_effect}

\begin{figure*}
\centering
\includegraphics[trim=000mm 000mm 000mm 000mm, clip=true, width=1.0\linewidth]{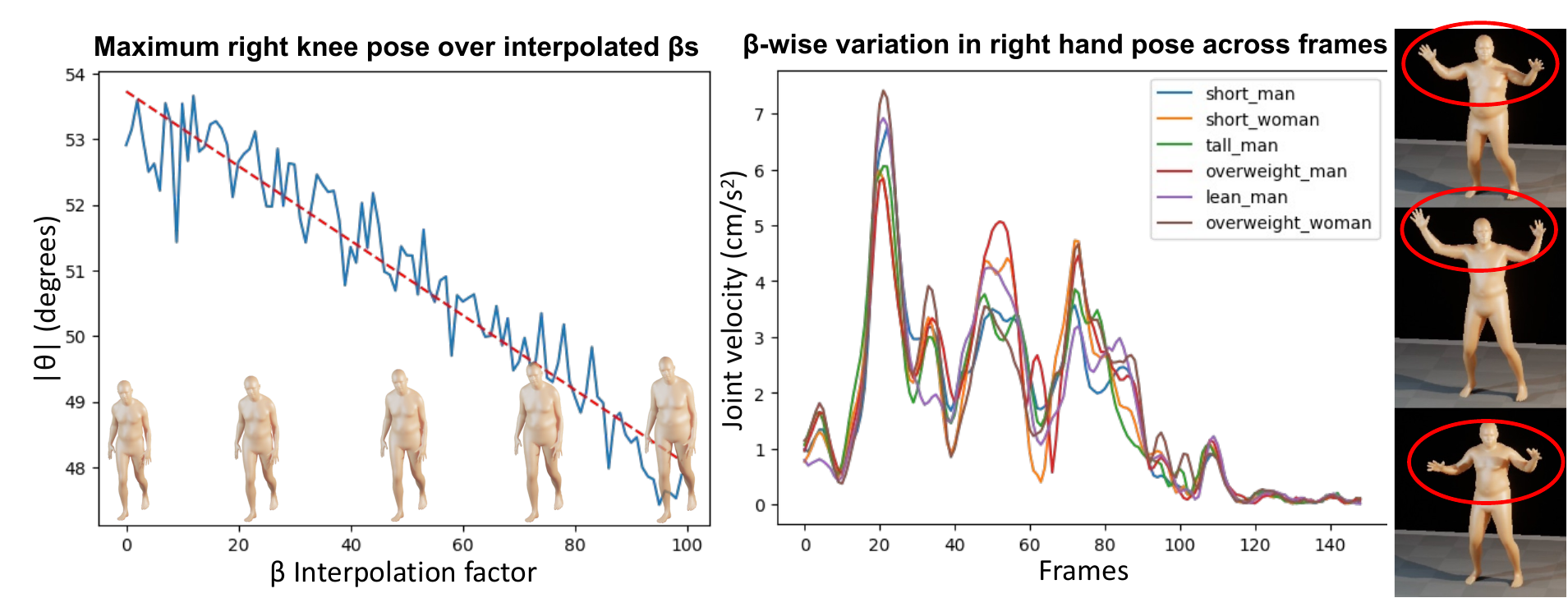}
\caption{Effect of body shape across (left) interpolated $\beta$ parameters, (center) 150 frames for 6 different identities and (right) different identities for the same jumping jack frame  \faSearch~~\textbf{Zoom in}.}
\label{figure:rebuttal:body_shape_effect}
\end{figure*}

\camready{
In \cref{figure:rebuttal:body_shape_effect} (left), we assess the diversity of \nameMethod generated motions across 100 $\beta$ parameters obtained by interpolating between a short male and a tall male body. We report the maximum right knee joint angle ($|\theta|$) for the same walk sequence shown in the \supmat (SM) video (05:34). The graph illustrates that taller people bend their knees less for the same walking motion, indicating body parameters affect movement. Similarly, in \cref{figure:rebuttal:body_shape_effect}~(center), we plot the right hand joint velocity across six different identities in the same walk sequence~[SM video (05:34)]. The joint velocities differ across subjects in corresponding frames implying diversity induced by body shape variation. In \cref{figure:rebuttal:body_shape_effect} (right), we qualitatively show the same frame of the jumping jack sequence~[SM video (05:47)] where the different arm positions indicate motion diversity.} 

\section{Additional Qualitative Results}
\label{supmat:additional_qualitative}

\begin{figure*}[ht]
\centering
\includegraphics[width=\linewidth]{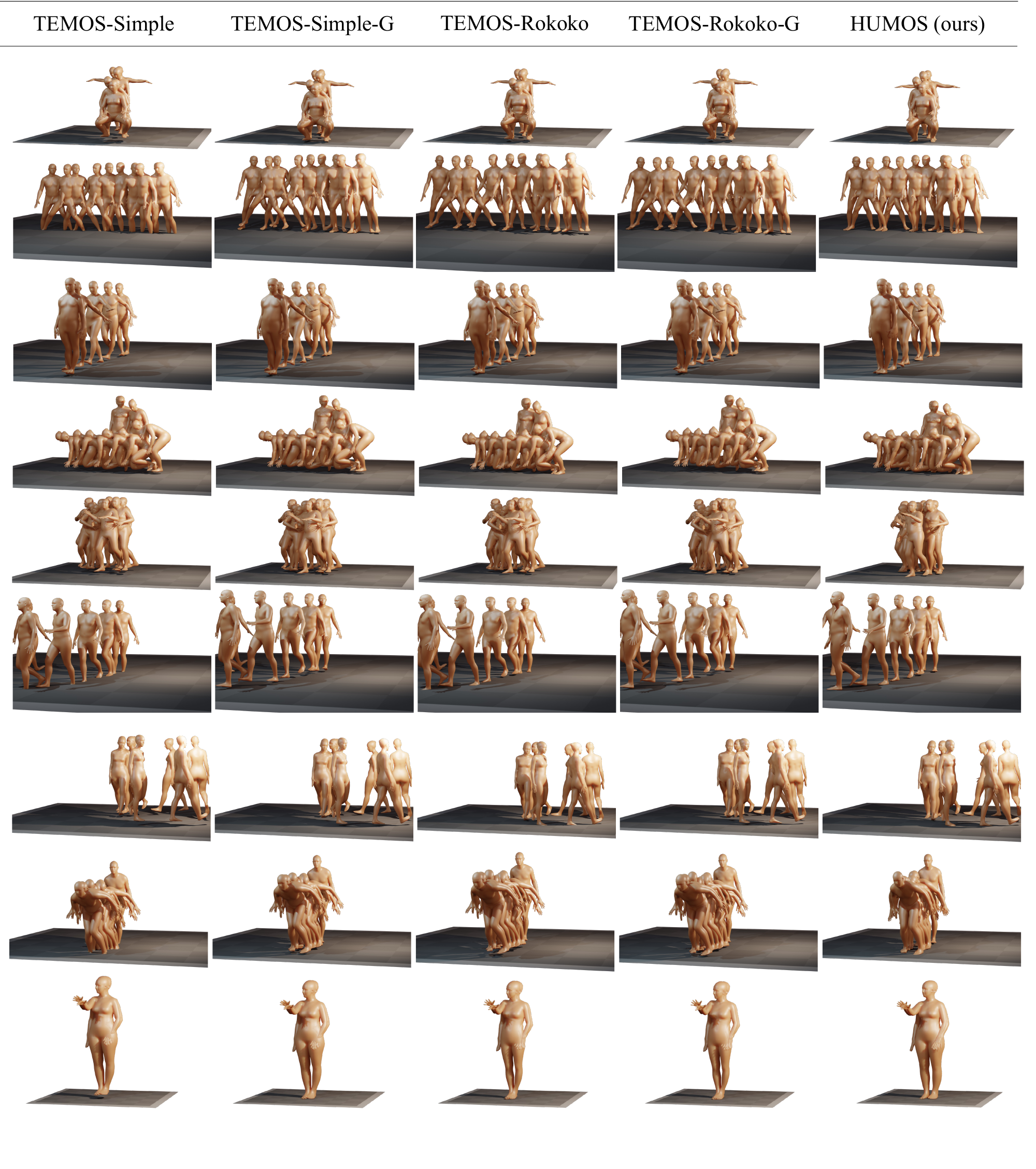}
\caption{Additional qualitative comparison of shape-conditioned motion generation. Each row represents generations across different methods for a unique body shape and gender. The difference in quality between methods is particularly evident in their interaction with the ground. \faSearch~\textbf{Zoom in}.} 
\label{figure:tipman:supmat:qualitative_comparison2}
\end{figure*}

We include additional comparisons with baselines in \cref{figure:tipman:supmat:qualitative_comparison2}. For video results, we recommend watching the  \textbf{supplementary video}. 

\section{Additional Ablations}
\label{supmat:latent_ablation}

In \cref{tab:latent_ablation}, we conduct additional ablations to analyze the effect of latent embedding losses, $\mathcal{L}_E$ and  $\mathcal{L}_{KL}$. We take the HUMOS model and successively remove the two loss terms individually. On ablating $\mathcal{L}_E$ during training, we observe a small improvement in ground penetration and float. However, the skate and dynamic stability metrics worsen. While the effect of $\mathcal{L}_E$ is minimal in terms of metrics, we empirically note faster and stable convergence when using it during training. $\mathcal{L}_{KL}$ also results in a slight improvement in physics metrics at the cost of dynamic stability. A significant advantage, however, of using $\mathcal{L}_{KL}$ is that it adds structure to the shape-agnostic latent space, making realistic motion generation easier.

\begin{table*}[!h]
\centering
\caption{Ablation study for latent embedding losses, $\mathcal{L}_E$ and $\mathcal{L}_{KL}$}
\label{tab:latent_ablation}
\begin{adjustbox}{width=1\textwidth}
\setlength{\tabcolsep}{4pt}
\begin{tabular}{@{}l|c|c|c|c|c@{}}
\toprule
\rule{0pt}{1.1EM}
       \textbf{Method} & 
       \multicolumn{1}{c|}{\textbf{Penetrate (cm)} $\downarrow$}                                & 
       \multicolumn{1}{c|}{\textbf{Float (cm)} $\downarrow$}                                    &
       \multicolumn{1}{c|}{\textbf{Skate (\%)} $\downarrow$}                                    &
       \multicolumn{1}{c|}{\textbf{Dyn. Stability (\%)}$\uparrow$}   &                 
       \multicolumn{1}{c}{\textbf{BoS Dist (cm)} $\downarrow$} \\
\midrule
\rule{0pt}{1.1EM}
HUMOS  & 1.23 & 1.04 & 7.37  & 71.9 & 14.62   \\ %
\midrule
\ HUMOS - $\mathcal{L}_{E}$  & 1.20 & 0.98 & 9.3 & 71.0 & 15.01   \\ %
\ HUMOS - $\mathcal{L}_{KL}$ & 1.14 & 0.93 & 6.96 & 71.05 & 15.21  \\ %
\hline
\end{tabular}
\end{adjustbox}
\end{table*}

\section{AMASS Shape Statistics}
\label{supmat:amass}

For training and evaluation, we use the AMASS dataset~\cite{AMASS_2019}. AMASS is a comprehensive collection of human motion data, unifying various optical marker-based motion capture datasets. This dataset stands out due to its extensive volume, containing over 50 hours of motion data from 480 unique subjects, encompassing more than 11,000 distinct motions. Among the 480 unique subjects, we have 274 male and 206 female subjects. To understand the diversity of body shapes included in AMASS, in \cref{figure:tipman:supmat:shape_mean_std}, we plot the mean and standard deviation of each principal component for the AMASS beta parameters. Following prior work, we use the first 10 shape principal components to represent body shape.

\section{Additional Implementation Details.}
\label{supmat:additional_implementation_details}

\begin{figure}[htbp]
\centering
\includegraphics[width=\textwidth]{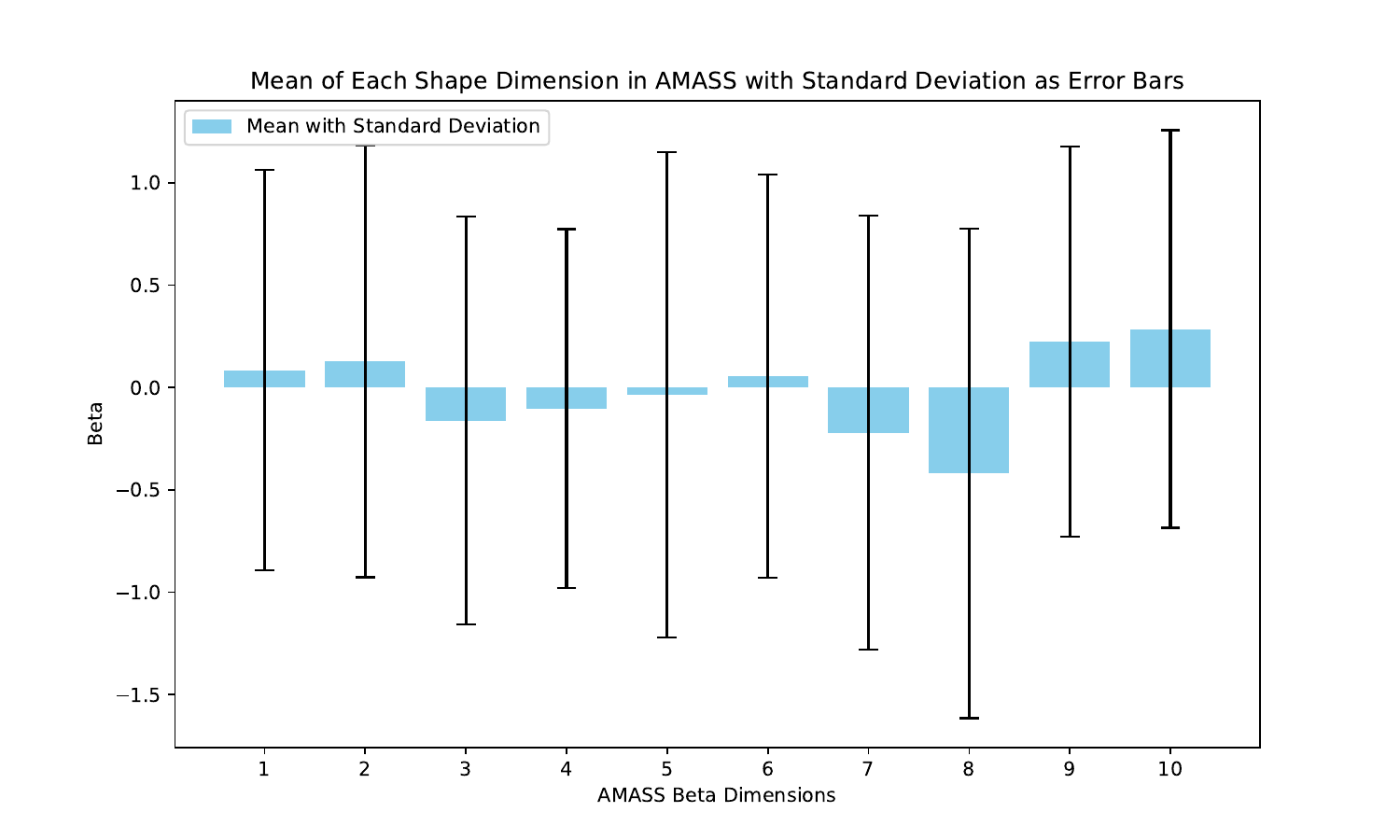}
\caption{Mean and standard deviation of the first 10 betas parameters in AMASS. This represents the diversity in body shapes.} 
\label{figure:tipman:supmat:shape_mean_std}
\bigskip
\includegraphics[width=\textwidth]{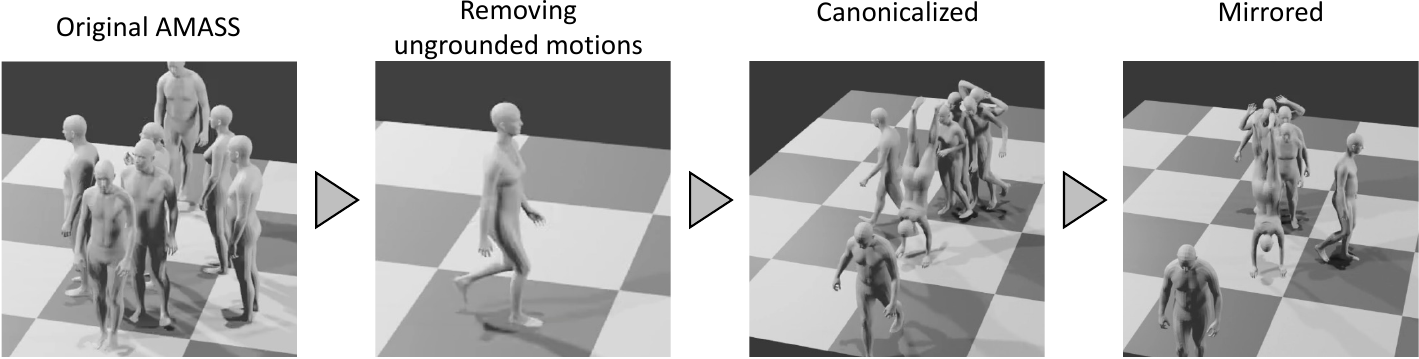}
\caption{We process the raw data from AMASS by 1) removing unsupported physically implausible motions \eg walking up the stairs 2) canonicalizing all motions to start facing the same direction at origin and 3) mirroring the pose and root translation to augment data} 
\label{figure:tipman:supmat:step_by_step}
\end{figure}

\noindent\textbf{Data processing.} AMASS captures diverse human motions performed by real participants. Therefore, motions in AMASS start at arbitrary locations and facing directions. AMASS also includes motions where the person is supported by objects such as chairs, stairs or raised platforms. Only the human is captured in such sequences, and given the lack of a supporting object, these motions are physically implausible. These sequence, along with arbitrary start locations and facing directions, add unnecessary ambiguity and make the raw AMASS data unsuitable for neural network training. To prevent this, we process the raw AMASS data by removing all sequences where the lowest vertex in at least 5 frames is higher than $0.25$m from the ground. Next, as described in the main text, we canonicalize all sequences to start at the origin with the same facing direction. To augment our training data, we mirror the pose parameters and global root translation from left-to-right and vice-versa, effectively doubling the training data. \Cref{figure:tipman:supmat:step_by_step} shows the effect of each step in our data processing pipeline.    

\noindent\textbf{Motion representation.} The SMPL body model parameterizes the human body into body pose, shape and global root translation. The SMPL body pose is represented as parent-relative rotations in the axis-angle format. For our motion representation, we follow NeMF~\cite{he2022nemf} and convert the parent-relative joint rotations to global root-relative rotations in 6d format~\cite{zhou20196d}. This helps with convergence and produces better performance than using the SMPL parameters directly. We also experiment with using deltas in joint rotations and global translation in our motion representation. We empirically observe worse performance in this setting due to the propagation of errors in the integration step when recovering the joint rotations and translation from the predicted deltas.

\subsection{Perceptual Study}

\begin{figure}[ht]
\centering
\includegraphics[width=\linewidth]{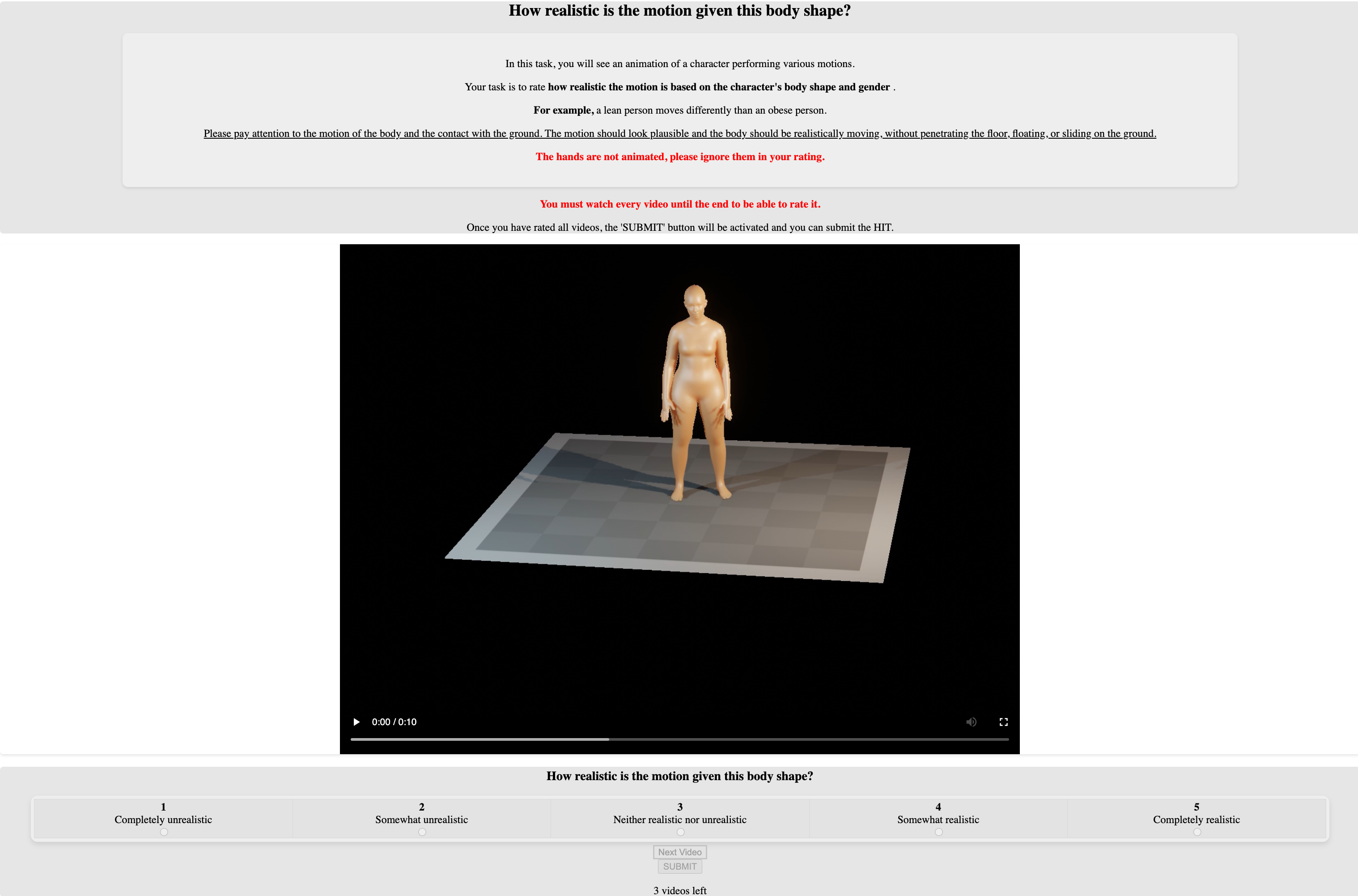}
\caption{Layout of the perceptual study.} 
\label{figure:tipman:supmat:perceptual}
\end{figure}

We show the layout for our perceptual study in \cref{figure:tipman:supmat:perceptual}. We randomly sample sequence generations from our methods and baselines and every video is rated by 25 participants on Amazon Mechanical Turk. We ensure quality in ratings by adding two ground-truth videos and two catch-trial videos per worker with extreme ground penetrations or floating sequences. Additionally, every participant is shown 5 \emph{warming-up} sequences at the start of their annotation task which we discard. This allows the participant to get a sense of the task before they can reliably rate the generated motions. We report average ratings across all participants who qualify the quality checks. 

\camready{To test statistical significance, we performed one-way ANOVA tests, yielding a significant p-value of $1.5\times e^{-10}$. Tukey's HSD statistical test indicates that our method has statistically significant differences with TEMOS-Rokoko (mean diff = 0.386, $p<0.001$) and TEMOS-Rokoko-G (mean diff = 0.447, $p<0.001$).}

\end{document}